\documentclass[10pt,twocolumn]{article}
\usepackage[margin=.82in,columnsep=.27in]{geometry}
\usepackage[T1]{fontenc}
\usepackage[utf8]{inputenc}
\usepackage{amsmath,amssymb,graphicx}
\usepackage{xurl}
\usepackage[unicode,colorlinks=true,linkcolor=blue,urlcolor=blue,citecolor=blue]{hyperref}
\usepackage{microtype}
\usepackage{caption}
\usepackage{balance}
\begin{document}
\title{Neural Language Models Learn the Contextual Distributions of Dependency Structures: a statistical learning theory to compositionality}
\author{Wang Bojun$^{1,*}$, Junjie Chen$^{2}$, Holly Jenkins$^{1}$, and Elizabeth Wonnacott$^{1}$\\[.5ex]
\small $^{1}$Department of Education, University of Oxford, Oxford, United Kingdom\\
\small $^{2}$Department of Statistics, University of Oxford, Oxford, United Kingdom\\
\small $^{*}$Correspondence: \href{mailto:wolf7335@ox.ac.uk}{wolf7335@ox.ac.uk}}
\date{}
\maketitle
\begin{abstract}
It is unclear how Neural Language Models (NLMs) acquire the structural meaning encoded by grammatical structures that is independent of lexical semantics. We propose a statistical learning process in which learned dependency structures themselves become new distributional units for subsequent statistical learning. Under this account, once a dependency structure is acquired, the model tracks its contextual distributions. These contextual features reflect the semantic properties of a composite structure. To test this hypothesis, we design a synthetic grammar in which each grammatical structure has distinct contextual distributions that cannot be recovered from the distributional statistics of their component tokens alone. We train a series of BERT-style masked language models on this grammar and examine their developmental trajectory. The results show that models can successfully learn the contextual distributions of composite dependency structures even though they cannot be inferred from token statistics alone. Developmental analysis further reveals a clear developmental trajectory. The learning of the dependency relations that define a grammatical structure consistently precedes the learning of its contextual features. These findings suggest that statistical learning in NLMs is not merely the accumulation of token co-occurrence statistics, but a process in which learned dependency structures become new units of distributional learning. We argue that this process provides a statistical-learning account of how NLMs solve the compositionality problem in language. Finally, we discuss the possibility that this statistical learning process provides an explanatory theory on how language cognition could emerge from pure distributional statistics.

\end{abstract}
\section{Introduction}
The cognitive science of NLMs endeavor to explain the language cognition and emergence of reasoning ability in NLMs. A myriad of research is dedicated on interpreting the semantic cognition (Marro et al., 2025; Vulic et al., 2020; Petersen \& Potts, 2023; Hardy et al., 2023; Hofmann et al., 2025), syntactic representation (Ahuja et al., 2024; Futrell et al., 2019; Hewitt \& Manning, 2019; Kim \& Smolensky, 2021; Lakretz et al., 2021; Murty et al., 2023; V\'{a}zquez Mart\'{i}nez et al., 2023; Wilcox et al., 2022; Futrell \& Mahowald, 2025; Li \& Liu, 2025; Linzen et al., 2016; Wei et al., 2021; Weissweiler et al., 2023a, 2023b), and logic deduction (Huang \& Chang, 2023; Mondorf \& Plank, 2024; Patil \& Jadon, 2025) in these models. One under-explored aspect of NLM cognition is statistical learning (Kallens et al., 2023; Kapatsinski, 2026; Bojun et al., 2026). That is, how the knowledge about these traditional concepts in human cognitive science such as syntax and semantics is constructed by simply tracking the distributional statistics in the input.

In human cognitive science, statistical learning research examines how human learners track statistical patterns in the linguistic input and how language cognition could be constructed by these learning mechanisms (Smith, 1969; Pelucchi et al., 2009; Saffran, 2001, 2020; Saffran et al., 1996; Brown et al., 2022; Lany \& Saffran, 2010, 2011; Mintz, 2002; Reeder et al., 2013, 2017; Morgan \& Newport, 1981; Thompson \& Newport, 2007; Wonnacott et al., 2017; Isbilen \& Christiansen, 2022; Misyak et al., 2009; Perek \& Goldberg, 2015, 2017; Samara et al., 2025). These studies design artificial languages that mirror certain statistical patterns in natural languages. By training humans on these artificial languages, researchers have gained lots of insight on how humans tease the intertwined statistical patterns in input and store these patterns as part of the linguistic knowledge. This constitutes part of the transition from representation to learning in cognitive science. (Perek, 2015; Pinker, 1989; Romberg \& Saffran, 2010; Saffran, 2020; Tomasello, 2003, 2007; Wonnacott, 2013; Kallens et al., 2023). Rather than looking into the static representation forms alone, the focus shifted to the process of how a representation is dynamically constructed.

Statistical learning constitutes an essential component for the cognitive science of NLMs because traditional concepts in cognitive science such as syntax, semantics and reasoning are all fundamentally statistical dependencies for NLMs (Kallens et al., 2023). The representation of NLMs contains only various kinds of statistical dependencies acquired through tracking distributional statistics in the input. Functional concepts are all reflections of these statistical dependencies. One major task in the cognitive science of NLMs is thus explaining the statistical nature of these traditional cognitive science concepts in the representation of NLMs.

The current study aims to provide a statistical learning theory to the compositionality problem in NLM language representation (Lake \& Baroni, 2018; Hupkes et al., 2020; Kim \& Linzen, 2020). Compositionality problem concerns how composite semantics is derived from the component semantics. For example, how the semantics of a sentences is derived from the lexical semantics of the words in the sentence. As the idiomatic expressions in (1), the semantics of the expression contains a conditioning relation which distinguishes ``the taller, the stronger'' from ``the stronger, the taller''. This conditioning structure is not explicitly reflected in any component lexical semantics (Culicover \& Jackendoff, 1999; Nunberg et al., 1994; Croft \& Cruse, 2004; Hilpert, 2019). Thus the composite semantics cannot be fully derived from lexical semantics alone. Another example is the linking problem in (2). In (2a), knowing the lexical semantics denoted by ``kill'' and the referent for ``Mary'' and ``John'' is not sufficient for deriving the composite semantics of the whole utterance. It would be confusing that whether Mary is the killer or John is the killer in this killing event. The representation of language has to contain a special component that specifies the mapping from syntactic elements to semantic elements (Levin \& Rappaport Hovav, 2005; Baker, 1997; Dowty, 1991). In this case, this special component needs to specify that the constituent before the predicate is the agent while the constituent after the predicate is the patient. In other words, representation of language needs to contain an additional component that guides the composition of component semantics in forming composite semantics.

(1) a. The more, the better.

b. the taller, the stronger

c. the more you read, the cleverer you get.

(2) a. Mary killed John.

b. Dad kicked the kangaroo.

c. Derek punched the boy.

The solution to the compositionality problem in human language representation is postulating a structural semantics corresponding to each grammatical structure as the additional component (Goldberg, 2019; Croft, 2001; Langacker, 2009; Levin, 2015). In this representation form, each surface grammatical structure is directly associated with a structural semantic schema. This form-meaning correspondence specifies not only the mapping from a grammatical structure to its semantic schema, but also the mapping from syntactic elements to the corresponding semantic elements. For example, the ditransitive structure in (3) is associated with the event structure in (4), the mapping from syntactic relations to semantic relations is also illustrated in (4). This representation form also applies to more idiomatic grammatical structures. For example, the ``the adj-er, the adj-er'' structure in (1) simply corresponds to a semantic schema that specifies the second comparative is conditioned on the first one.

(3) ditransitive grammatical structure:

\begin{center}
\includegraphics[width=.85\linewidth,height=.40\textheight,keepaspectratio]{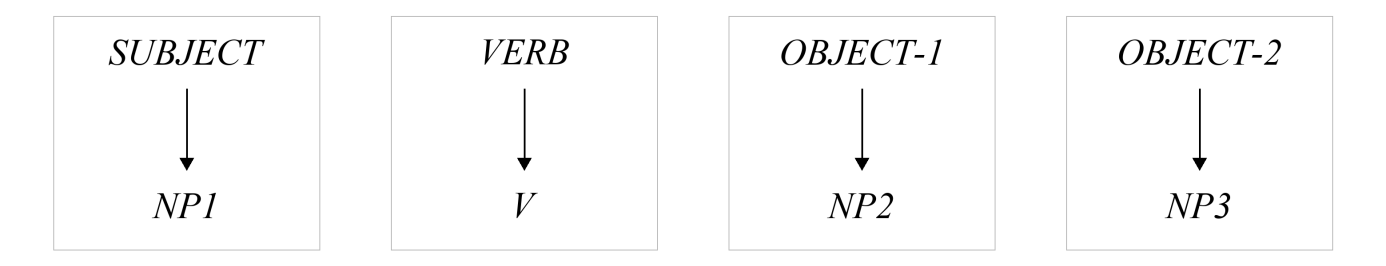}
\end{center}
(4) structural meaning of ditransitive structure

\begin{center}
\includegraphics[width=.85\linewidth,height=.40\textheight,keepaspectratio]{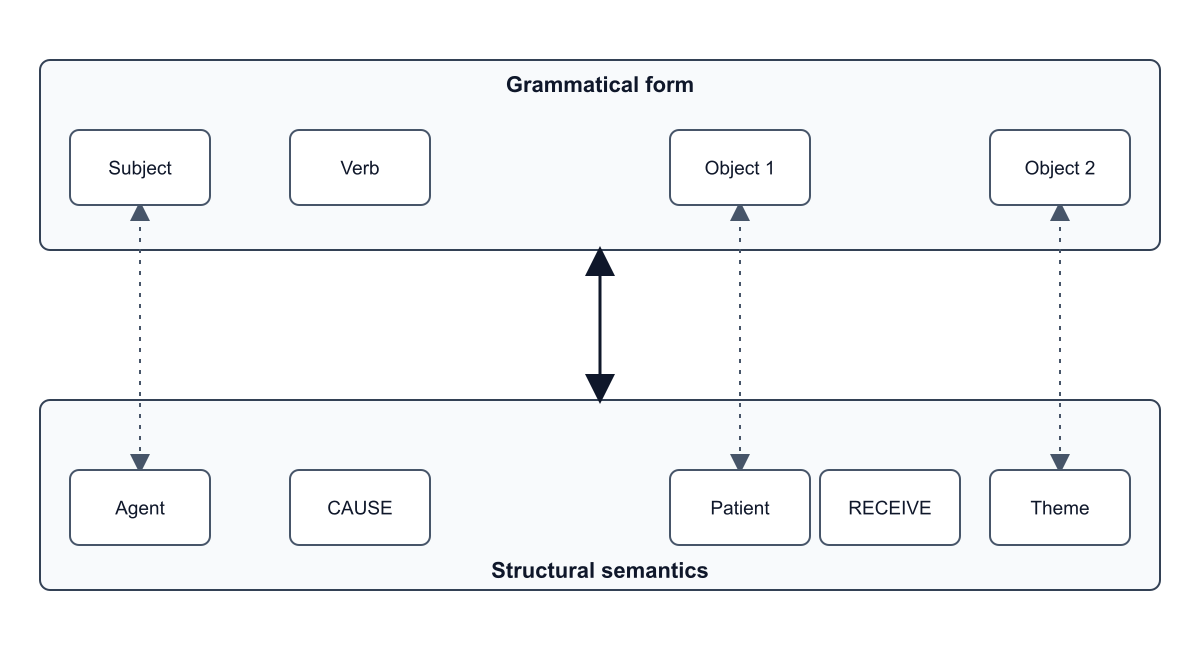}
\end{center}
(5) e.g.

a. John gave Mary a book.

b. Derek sent Viona a box.

c. Alice threw the dog a ball.

For human learners, these correspondences between grammatical structures and structural semantics are acquired through the generalizations on the relations between input utterances and their real discourse context (Tomasello, 2003; Perek \& Goldberg, 2017; Croft, 2003). For example, by observing sentences in (5) repetitively in real discourse context, learners could associate the ditransitive syntactic structure to the transfer of possession event structure. Further, learners could figure out the constituent before the verb is always the giver and the first constituent after the verb is always the recipient. The representation of language then consists of a huge collection of such generalizations on form-meaning correspondences. These form-meaning correspondences serve as templates when deriving the composite semantics from its lexical components, since they also entail the linking from syntactic elements in the grammatical structure to the semantic elements in the structural semantic schema.

In this study we ask the question how NLMs solve the compositionalality problem. NLMs cannot learn such correspondences by attending to the semantic environment of input utterances like humans. For NLMs, the only way of acquiring semantic information of an element is tracking its distributions against other elements in its context.

We propose that there is a statistical learning process that tracks the distributional profile of composite dependency structures in the input. We hypothesize that when a composite dependency structure is learned, the model would treat this composite structure as a distributional unit and generalize on its contextual features. That means tracking the dependency relations between this composite structure and other distributional units in its context. These contextual features of a composite dependency structure reflect its semantic properties and linking rules.

We define internal dependencies are the dependency relations between token categories that constitute a composite dependency structure. Surface grammatical structures are instances of such composite structures. The syntactic relations that define a grammatical structure correspond to such internal dependency relations. We then define external dependencies as the dependency relations between a composite structure and other distributional units in its context. Collectively, these external dependency relations constitute an external distributional profile, which describes the distributional propensity of the composite structure. This external distributional profile represents the properties intrinsic to the composite structure, thus cannot be captured by the internal dependencies which contain no contextual information for the composite structure.

This theory makes an assumption that surface grammatical structures have distinct distributional profiles in large corpora. For example, ditransitive structure encoding ``transfer of possession'' meaning should have a very different distributional profile with the grammatical structure encoding ``voluntary motion'' events. Similarly, the grammatical structure encoding causation should have very different distributional profiles with the grammatical structures encoding the meaning ``a property modifying a thing''. Even for grammatical structures encoding similar event semantics, the distributional profile should diverge as long as there is even slight difference in function. For example, even though ditransitives in (5) and datives in (6) both encode ``transfer of possession'' meaning, they have functional differences on discourse context. Both experimental studies and corpus studies observed that they are used in different contexts. For example, the context of ditransitives usually contains a definite recipient participant, which is previously given in discourse; the context of datives instead is inclined to contain an indefinite and newly introduced recipient participant (Rappaport Hovav \& Levin, 2008; Goldberg, 2019; Bresnan et al., 2007; Theijssen et al., 2013). These nuanced functional differences are also sources of contextual distributional difference in large corpora.

(6) a. Mary gave a book to John.

b. Derek sent a box to Viona.

c. Alice threw a ball to the dog.

In this study, we test whether NLMs form external dependencies for composite dependency structures and how the learning path in this process looks like. We design a synthetic grammar where grammatical structures are defined by internal dependencies, and these grammatical structures have unique external distributional profiles. We train a series of encoder-only Transformer models on this synthetic grammar and conduct two analysis: (a) clustering of vectors based on internal and external dependencies; (b) probability mass in sequence completion. Finally, we scale the models to demonstrate the generalizability of the findings.

The synthetic grammar:

The synthetic grammar consists of 5 categories of tokens, labeled as A, B, M, N, U. Among them, A, B, M and N categories each contain k (k=20) distinct tokens. U category contains 100 distinct tokens.

There are two grammatical structures in this synthetic grammar, BAA and ABA. These two trigrams are composite dependency structures defined by the dependency relations between A and B categories. M and N are contextual categories of BAA and ABA structures. M only occurs in the context of BAA structure and N only occurs in the ABA structure. Thus M tokens are the contextual features of BAA composite structure; N tokens are the contextual features of ABA structure. U tokens are the noise. Their distribution is totally random.

The core property of the synthetic grammar is that M and N categories cannot be distinguished by their token-level distributional statistics. Both BAA and ABA structures contain two A tokens and one B token. Consequently, M and N categories cannot be distinguished by their distributions against A and B tokens. The distributional difference between M and N categories only emerge when the composite dependency structures BAA and ABA are represented as distributional units. Thus a model would only distinguish M and N categories if it attends to the external distributional profile of the composite structures.

Below is the dataset generation procedure. The dataset consists of a core set and a compensation set. In the core set, we first generate all possible BAA-M and ABA-N combinations. This creates 2k4 possible combinations. We initialize a noise dataset that contains only randomly distributed U tokens. Each sequence contains seven U tokens, there are 2k4 such noise strings in the core set in total. Then for each string, we randomly select three connected positions and replace these three positions by a trigram of either BAA or ABA. Then for the four remaining positions, we randomly select one position to replace the original U token by a contextual token. If the trigram is BAA, then the contextual token is an M token; if the trigram is ABA, the contextual token is an N token. The 2k4 strings thus exhaust all possible BAA-M and ABA-N token combinations. Each specific BAA-M or ABA-N combination occurs precisely once in this dataset.

To this point, the core set introduces a negative dependency between M and N categories since they never co-occurs. We thus create a compensation set to offset this negative dependency to ensure M and N has no dependence in the whole dataset. We set the compensation set to contain 3k4 strings, thus the whole dataset contains 5k4 strings. These compensation strings is initialized to contain only seven randomly distributed U tokens. We insert M and N tokens in the compensation set to create controlled co-occurrence of M and N to offset the negative dependency as described below.

We define P(M) as the proportion of sequences containing an M token in the complete dataset of 5k4 sequences.~The same applies to P(N). Under independence, the joint probability must satisfy P(M,N)=P(M)P(N). In the core set, k4 sequences contain M and the other k4 contain N. Thus, relative to the complete dataset, the core set contribute k4/5k4=0.2 to P(M) and 0.2 to P(N), whereas it contributes 0 to P(M,N) because M and N never co-occur here. The compensation set is designed to remove this negative dependency. We set the whole dataset marginal probabilities to P(M)=P(N)=0.4, which requires P(M,N)=0.16. This indicates 16\% of all sequences contain both M and N token. For these sequences, we insert one random M token and one random N token. Because the core set already contributes 20\% sequences with M token and 20\% sequences with N token, the compensation set need to additionally contributes 4\% M-only sequences and 4\% N-only sequences to ensure the marginal probability P(M)=P(N)=0.4. For M-only sequences, we randomly select one M token and insert it to the noise base; for N-only sequences, we randomly select one N token and insert it to the noise base. The complete dataset therefore consists of 40\% core sequences, 16\% M-N co-occurrence sequences, 4\% M-only sequences, 4\% N-only sequences, and 36\% noise sequences containing only randomly distributed U tokens. This is illustrated in table 1.

\begin{table*}[!t]
\centering
\caption{Composition of the artificial-language dataset}
\label{tab:1}
\small
\begin{tabular}{p{.17\linewidth}p{.17\linewidth}p{.22\linewidth}p{.31\linewidth}}
\hline
Dataset subset & Proportion of complete dataset & Sequence type & Example seven-token sequence \\
\hline
Core set & 20\% & BAA with M & M  U  B  A  A  U  U \\
Core set & 20\% & ABA with N & U  A  B  A  U  U  N \\
Compensation set & 16\% & M and N co-occur & U  N  U  U  M  U  U \\
Compensation set & 4\% & M only & U  U  U  U  U  M  U \\
Compensation set & 4\% & N only & N  U  U  U  U  U  U \\
Compensation set & 36\% & Noise only & U  U  U  U  U  U  U \\
\hline
\end{tabular}
\end{table*}
In the examples, each capital letter is a category of tokens. Token positions are randomized independently in every sequence and therefore are not fixed. The composition gives P(M) = P(N) = 0.40 and P(M,N) = 0.16

The core set and compensation set are combined and shuffled as a whole dataset. In the final dataset, there is no dependency between M and N categories. The only potentially learnable dependency relations in this dataset are the internal dependencies that define BAA and ABA structures, and the external dependencies BAA-M and ABA-N.

We then create a control dataset. Based on the existing experimental dataset, we replace all BAA trigrams by ABA trigrams. The dataset then contains only one grammatical structure ABA. Both M and N tokens are contextual tokens of ABA structure. In this case, there is no distributional difference between M and N categories at all. We expect to see that for the controlled dataset, models cannot distinguish M and N category. Instead, models would believe M and N tokens together form one category, which constitutes the external distributional profile of the ABA structure. But for the experimental dataset, we expect to see that models would clearly distinguish M and N categories since they are the contextual categories of different composite structures.

(7)  a. Experimental set:  BAA--M;  ABA--N\\    b. Control set:       ABA--M;  ABA--N

In training, we randomly split the whole dataset into 90\% training set and 10\% validation set. This prevents the model from seeing all BAA-M and ABA-N combinations. To capture the unobserved data, the model has to generalize on the distributional propensity.

\section{Results}
\subsection{Vector analysis}
We train an encoder-only transformer model on this synthetic grammar using Masked Language Modeling (MLM). The model has 8 Transformer layers, 8 Attention heads, and an embedding dimension of 32 (8L-8H-32D). During training, we randomly mask two tokens in each sequence. Training loss is calculated on these two masked positions. The model is trained for 100,000 iterations and is saved after each 200 iterations. This gives us 501 model states in a developmental sequence. For the vector analysis, we extract the Value vectors from all attention heads in the first Transformer layer and concatenated all head-specific vectors into one. This is done for all tokens in the vocabulary. Vectors are extracted from all saved learning stages, so that we could look into how the clustering of these vectors changes in the course of development. After extracting all the vectors, an auto-encoder is trained to reduce the dimensions of vectors to 2D.

Another encoder-only Transformer model is trained on the control language following the same MLM strategy. This model shares the same architecture and hyper-parameters with the model trained on base language. This model is trained for 1,500,000 iterations, which is 15 times more than the base model. The vector extraction and dimensionality reduction technique are also consistent.

Figure 1 illustrates the clustering change of the base model throughout the learning path. At the initial stage of learning, there is no clustering. After 4,000 iterations training, the model starts to distinguish A and B categories in vector representation. This is the sign for learning internal dependencies since these two categories are defined by their dependency relations in these two grammatical structures. At this moment, the model fails to distinguish M and N tokens. It makes M and N tokens as a single cluster. After 39,200 iterations of training, the model successfully categorized M and N tokens into two distinct clusters. A motion chart is created to show the full learning path. \href{https://study1-trajectories-nx47.bojun-fl.chatgpt.site/model/8l8h32d-base.html}{Click here to view the motion chart.}

\begin{figure*}[!t]
\centering
\includegraphics[width=\textwidth,height=.48\textheight,keepaspectratio]{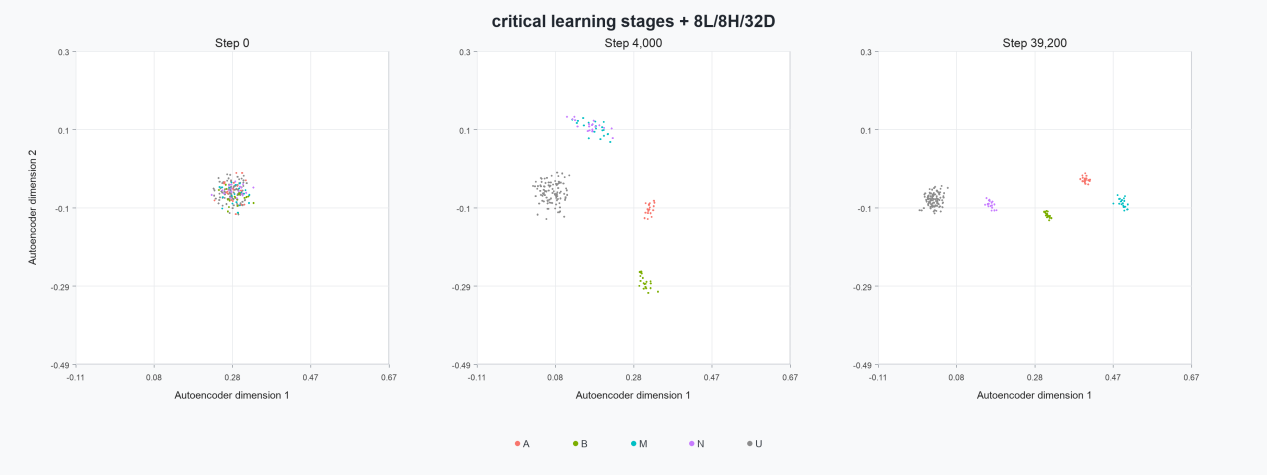}
\caption{Developmental change in the base-language model's first-layer V-vector representations: categories overlap at initialization, A and B separate by step 4,000, and M and N separation develops by step 39,200.}
\label{fig:1}
\end{figure*}
\begin{figure*}[!t]
\centering
\includegraphics[width=\textwidth,height=.48\textheight,keepaspectratio]{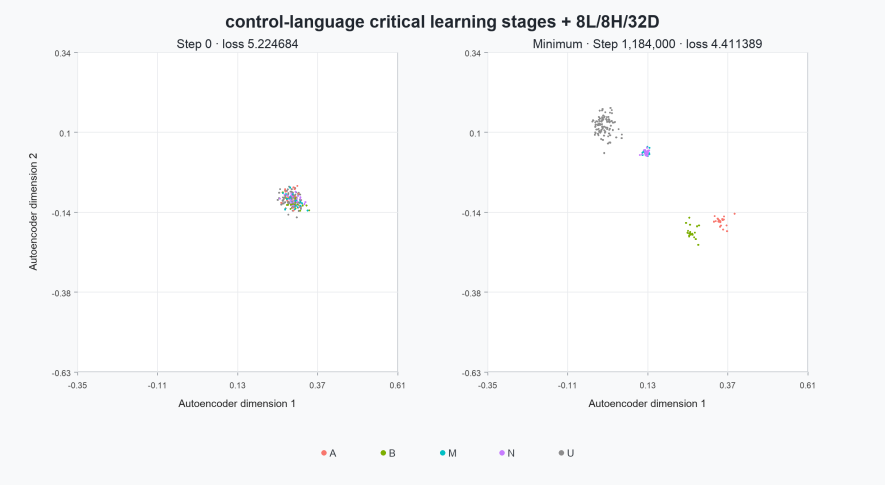}
\caption{First-layer V-vector projections for the control-language model at the initial and minimum-validation-loss checkpoint; A and B are separated, whereas M and N remain overlapping.}
\label{fig:2}
\end{figure*}
Figure 2 shows the clustering of the model trained on the control language at the initial state and the stage with lowest validation loss. \href{https://study1-trajectories-nx47.bojun-fl.chatgpt.site/model/8l8h32d-control.html}{Click here to view the full motion chart.} At minimum validation-loss stages, the model learned to categorize A and B tokens. But even though training length is prolonged 15 times, the model still fails to categorize M and N tokens in all checkpoints. Instead, the model forms a cluster that contains both M and N tokens. This indicates that the model is aware that M and N tokens are different from U tokens. This is expected because U tokens are totally randomly distributed, but M and N tokens are dependent on A and B tokens. The contrast between the experimental language model and the control language model proves that the categorization on the M and N tokens in the experimental language model is indeed based on the contextual features of BAA and ABA composite structures, rather than on lexical dependencies.

\subsection{Probability mass in sequence generating}
We create four sets of testing sequences. First two of them are designed to test the learning of internal dependencies, the last two are for external dependencies. All these strings are engineered from the strings in the original validation set, so none of them has been observed in training before. They are illustrated in (8) and (9). For the two sets of internal-dependency testing strings, we first find all strings containing BAA and ABA trigrams in the base validation set. We then replace all M and N tokens in these strings by random U tokens. So these strings only contain BAA and ABA trigrams at random positions. Then we mask the token in the middle position of these trigrams. This creates two kinds of trigram structures of B-MASK-A and A-MASK-A. They occur at random positions in each string. For the A-MASK-A testing strings, the target category is B; for the B-mask-A testing strings, the target category is A.

For the external-dependency testing, we again use strings in the original validation set as base. We first find all BAA and ABA strings. We then eliminate 4 kinds of strings: (a)strings starting with BAA trigrams, (b)strings ending with ABA trigrams (c) BAA strings where an M token immediately precedes the BAA trigram, (d)ABA strings where an N token immediately follows the ABA trigram. For the remaining strings, we preserve the M and N tokens as contextual cues. For BAA strings, we mask the final A position and the position before the B token. This creates a MASK-B-A-MASK structure. For the ABA strings, we mask the first A position and the position after the final A position. This also creates a MASK-B-A-MASK structure. If the context of the MASK-B-A-MASK structure contains an M token, the target of prediction is A category tokens in the second MASK position. If the context of MASK-B-A-MASK structure contains an N token, the target of prediction is A tokens in the first MASK position.

(8) example internal dependency testing

a. U-U-U-A--MASK--A-U

\(\to\) Target prediction: B category

b. U-U-B--MASK--A-U-U

\(\to\) Target prediction: A category

(9) example external dependency testing

a. M-U-MASK--B--A--MASK-U\\\(\to\) Target prediction: A category at the second MASK

b. U-MASK--B--A--MASK-N-U

\(\to\) Target prediction: A category at the first MASK

We conduct testing on each saved learning stage. In each stage, we randomly draw 512 testing strings of each type to do testing. For each testing string, we look into the probability distribution generated by the model on the target positions. We define probability mass as the sum of the probability for all tokens in the target category. For internal-dependency testing strings with A-MASK-A strings, we sum the probability of all B tokens at the masked position as the probability mass for B category; for testings with B-MASK-A strings, we sum the probability of all A tokens at the masked position. For external dependency testings with M tokens as contextual cues, we sum the probability of all A tokens at the second MASK position; for external-dependency testings with N tokens as contextual cues, we sum the probability of all A tokens at the first MASK position. If a dependency relation is learned, we expect to see the probability mass of the target category gradually approaches one in the learning path. That means the model is assigning almost all probability to tokens from the target category.

At each learning stage, we average the target category probability mass across 512 sampled testing strings within each probe set. Further, we train one hundred models with the same configuration but different seeds for initialization and data shuffling. The probability mass is then averaged across the one hundred models. The result is presented in Figure 3.

\begin{figure*}[!t]
\centering
\includegraphics[width=\textwidth,height=.48\textheight,keepaspectratio]{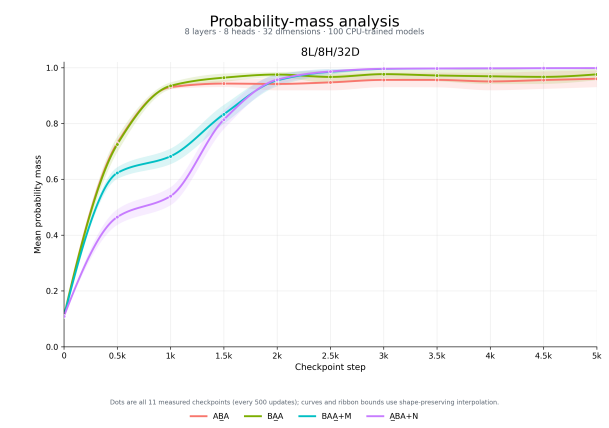}
\caption{Probability mass analysis through the learning path. Solid lines show the mean probability mass, ribbons show the 95\% confidence interval.}
\label{fig:3}
\end{figure*}
We add a control testing. For the same set of testing strings, we replace the context token M/N by a random U token. For the control testing, we calculate the difference between the probability mass of A category at the first MASK position and the second MASK position. For the original testing strings with M and N tokens, we calculate the difference between the probability mass of A category at the target position and the non-target position. We again do this analysis across 100 models with different randomization seeds and average.

\begin{figure*}[!t]
\centering
\includegraphics[width=\textwidth,height=.62\textheight,keepaspectratio]{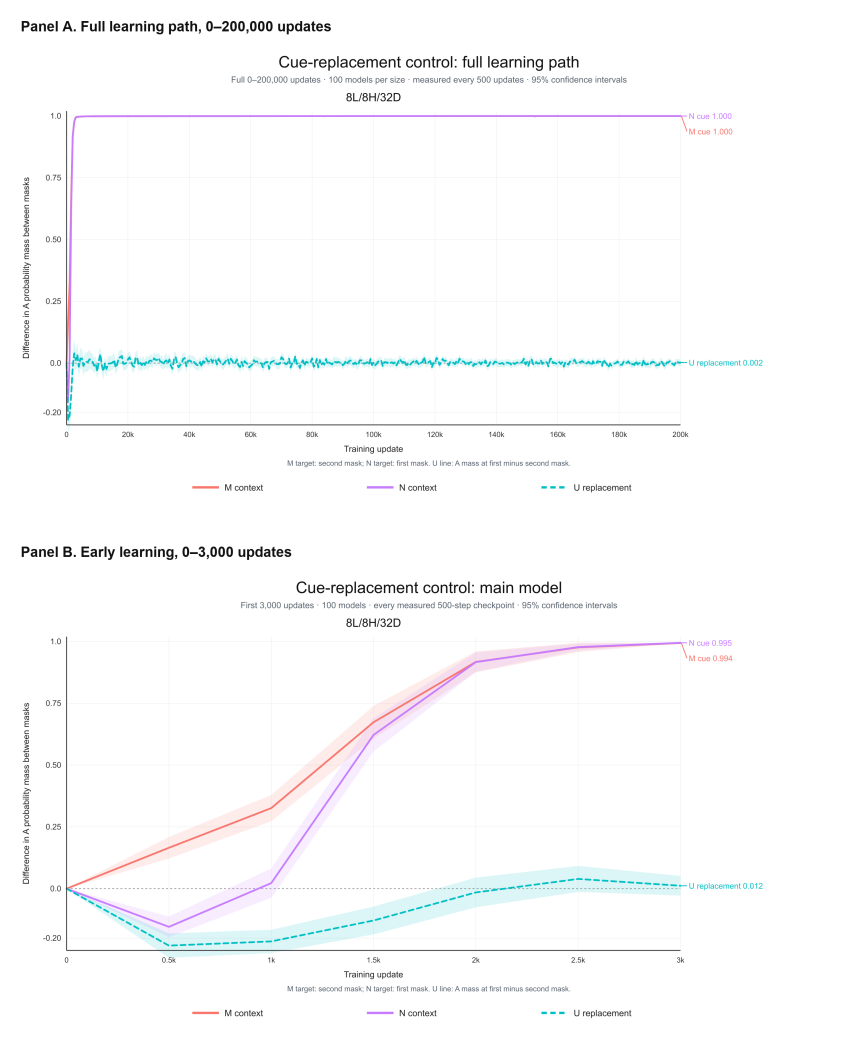}
\caption{Controlled probability-mass difference across 100 models for M, N, and random-U contexts. (Panel A) Full training path; (Panel B) close-up for the first 3,000 updates. Solid lines show model means; ribbons show 95\% confidence intervals.}
\label{fig:4}
\end{figure*}
The results suggest that NLMs indeed acquire external dependencies of composite dependency structures. The probability mass for all four sets of testing strings approached one in the end of learning. Note that these testing strings contain BAA-M and ABA-N combinations that never observed in training. Therefore, the only possible interpretation to the probability mass pattern in figure 3 is that the model formed productive categories defined by the distributional relations between the composite structures and their contextual tokens.

Further, the result suggests that the learning of internal dependency does not fully precede the learning of external dependency. The model does not first fully acquire the internal dependencies and then starts generalizing on the external dependencies. Instead, the model starts the learning of external dependencies at the time it first detects internal dependency relations. In other words, once the model found a distributional bias from the noise, the model would start tracking the contextual distribution of this distributional bias. Albeit, internal dependencies are generally learned earlier than external dependencies.

\subsection{Scaling the model}
We scale the model size to demonstrate that the observation is not subject to model size. Specifically, we compare three masked language models with increasing model capacity: 4L-4H-16D, 8L-8H-32D, and 8L-8H-64D. Here, L denotes the number of Transformer layers, H denotes the number of attention heads, and D denotes the hidden dimension.

For each architecture, we train 100 models with the same training configuration but different random seeds for model initialization and data shuffling. We apply the same probability-mass probes to every saved checkpoint of each trained model. For each checkpoint and each seed, we again randomly select 512 testing strings for each one of the four types. The probability mass is averaged across 512 sampled testing strings within each probe set. We then average this value across the one hundred seeds for each architecture. The result is illustrated in figure 5.

\begin{figure*}[!t]
\centering
\includegraphics[width=\textwidth,height=.48\textheight,keepaspectratio]{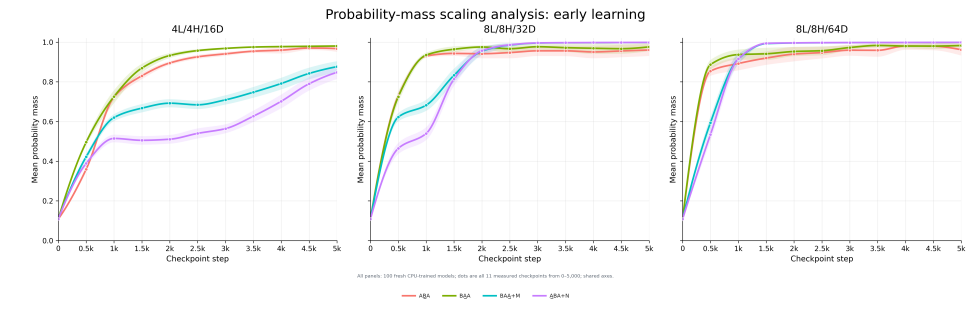}
\caption{Probability mass analysis for 4L-4H-16D models, 8L-8H-32D models, and 8L-8H-64D up to 5,000 steps.}
\label{fig:5}
\end{figure*}
We also include the control testing for the scaled models, in which we place the context token M/N in testing strings by a random U token. The procedure remains identical. The result is presented in figure 6.

\begin{figure*}[!t]
\centering
\includegraphics[width=\textwidth,height=.62\textheight,keepaspectratio]{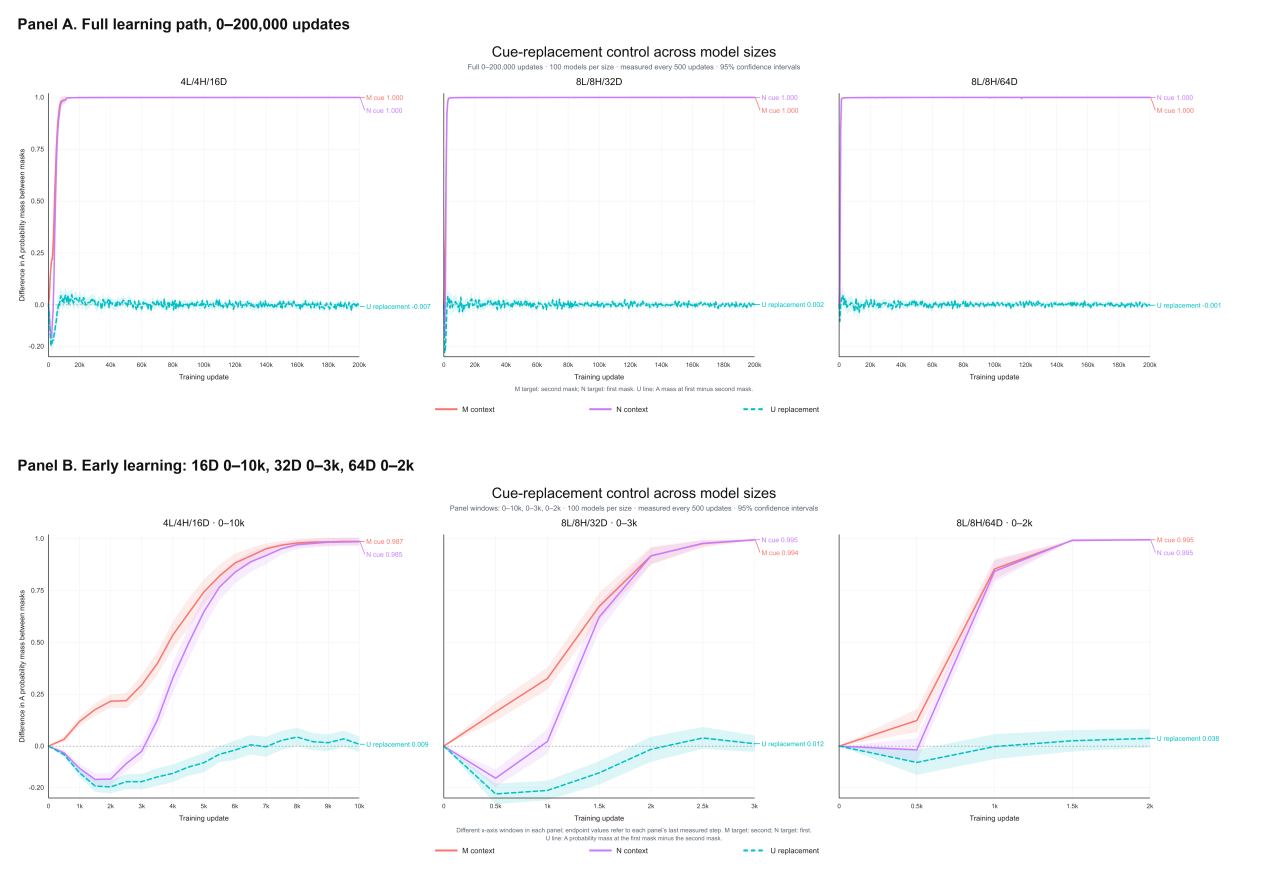}
\caption{(Panel A) Full training paths; (Panel B) early learning, zoomed to 10,000, 3,000, and 2,000 updates, respectively. Lines show model means; ribbons show 95\% confidence intervals.}
\label{fig:6}
\end{figure*}
For each of the two additional model configurations, we performed vector analysis on a model trained on the experimental language and an architecture-matched model trained on the control language. Figures 7 and 9 show critical learning stages from the experimental-language motion charts for the 4L--4H--16D and 8L--8H--64D models, respectively. View the full experimental-language motion charts: \href{https://study1-trajectories-nx47.bojun-fl.chatgpt.site/model/4l4h16d-base.html}{4L-4H-16D} and \href{https://study1-trajectories-nx47.bojun-fl.chatgpt.site/model/8l8h64d-base.html}{8L-8H-64D}. Figures 8 and 10 show the corresponding control-language models at initialization and at their minimum-validation-loss checkpoints. View the full control-language motion charts: \href{https://study1-trajectories-nx47.bojun-fl.chatgpt.site/model/4l4h16d-control.html}{4L-4H-16D} and \href{https://study1-trajectories-nx47.bojun-fl.chatgpt.site/model/8l8h64d-control.html}{8L-8H-64D}.

\begin{figure*}[!t]
\centering
\includegraphics[width=\textwidth,height=.48\textheight,keepaspectratio]{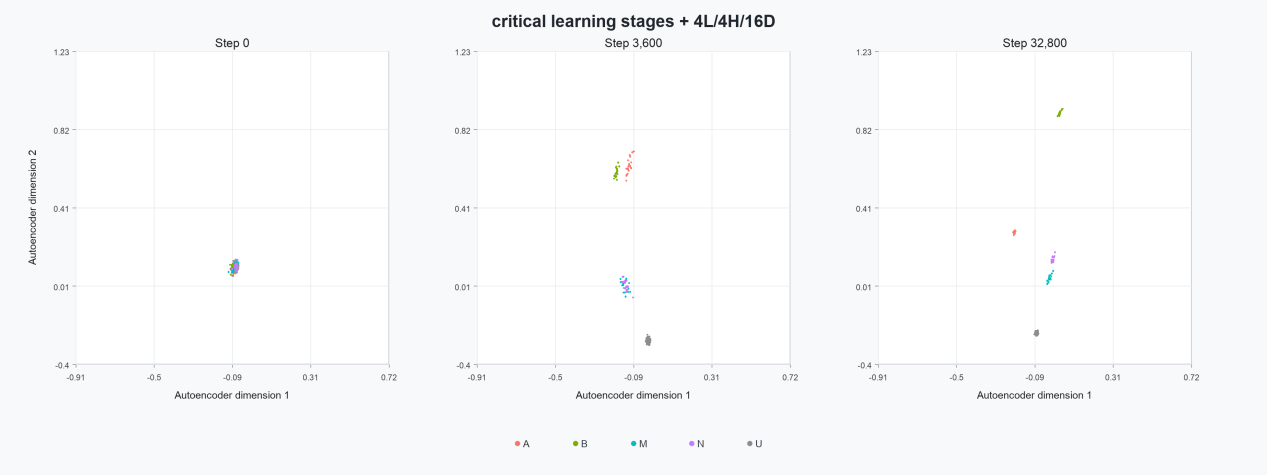}
\caption{Static first-layer V-vector trajectories of the 4L-4H-16D base-language model. The panels show training steps 0, 3,600, and 32,800.}
\label{fig:7}
\end{figure*}
\begin{figure*}[!t]
\centering
\includegraphics[width=\textwidth,height=.48\textheight,keepaspectratio]{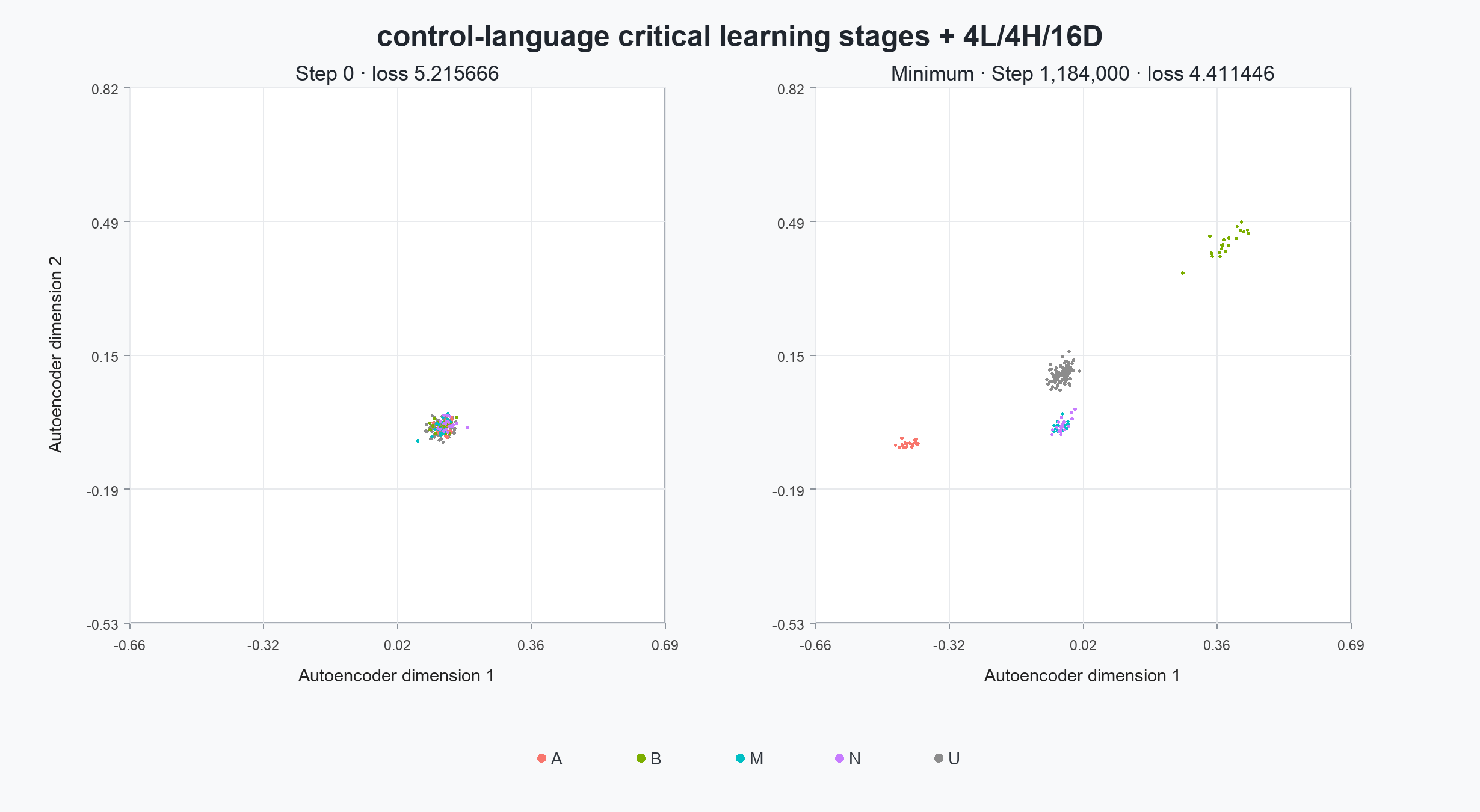}
\caption{4L-4H-16D model trained on the control language. Initial stage clustering and lowest validation-loss clustering.}
\label{fig:8}
\end{figure*}
\begin{figure*}[!t]
\centering
\includegraphics[width=\textwidth,height=.48\textheight,keepaspectratio]{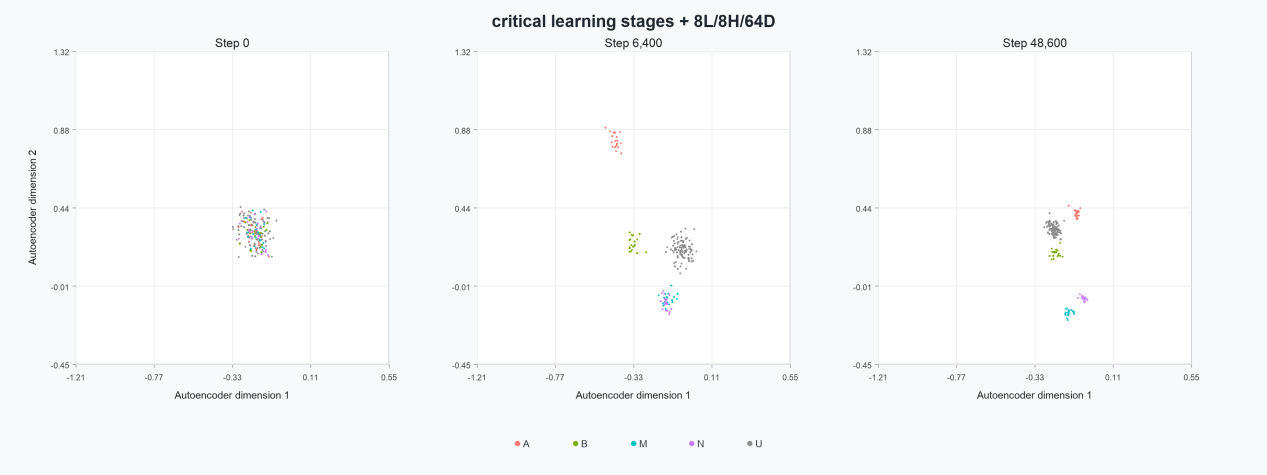}
\caption{Static first-layer V-vector trajectories of the 8L-8H-64D base-language model. The panels show training steps 0, 6,400 and 48,600.}
\label{fig:9}
\end{figure*}
\begin{figure*}[!t]
\centering
\includegraphics[width=\textwidth,height=.48\textheight,keepaspectratio]{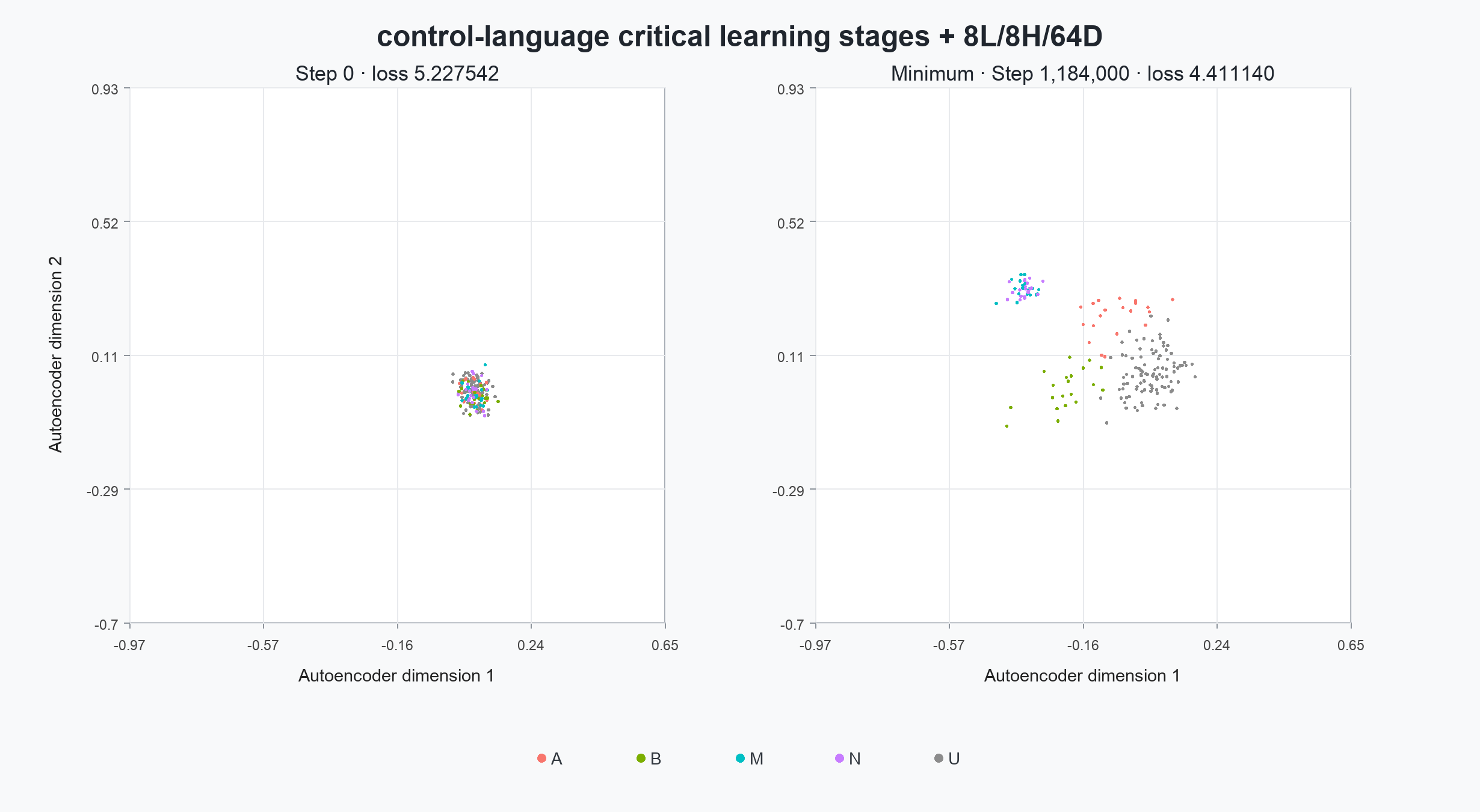}
\caption{8L-8H-64D model trained on the control language. Initial stage clustering and lowest validation-loss clustering.}
\label{fig:10}
\end{figure*}
These results suggest all findings for the 8L-8H-32D model could be replicated for the scaled models. The previous observations are not subject to model sizes.

\clearpage
\section{Discussion}
The observations in this study confirm the hypothesis that composite dependency structures constitute distributional units in the statistical learning of NLMs. In this process, NLMs form distributional profiles for dependency structures by tracking their contextual features. This statistical learning mechanism allows models to learn the properties that belong to the dependency relations, which cannot be inferred from the properties of component tokens.

The synthetic grammar in this study creates a radical scenario where two composite dependency structures are defined by the same set of tokens. This creates a clean statistical environment, in which contextual features of the two composite dependency structures are disentangled from token-level distributions. Natural language corpora are much more complex statistical environments. Distributional profiles of composite dependency structures may overlap with the distributional profiles of component tokens or token categories. It is possible that token x has a global repulsive relation to token y in the whole dataset, but the relation becomes locally attractive when y occurs in certain composite dependency structures. Further, it is also possible that the repulsive relation between x and y becomes locally stronger when y occurs in certain composite dependency structures. Further investigation is needed to establish how NLMs learn in these more complex statistical environments.

Another potential statistical structure that is not present in our synthetic grammar is the dependency relations between composite dependency structures. Since composite dependency structures are distributional units in the statistical learning of NLMs, the contextual features of a composite dependency structure are very likely to subsume its dependencies against other composite dependency structures. In other words, dependency structures might attract or repulse each other in large dataset. This suggests that the statistical environment in natural language corpora might be much more complex than merely token dependencies.

Further, the current study only demonstrates one level of the nesting between internal dependency relation and external dependency relation. Namely when a dependency relation is learned, it serves as an internal dependency and the model will learn its external distributions. A potential extension is recursive nesting: when an external dependency is learned, a larger composite dependency structure might be formed, where this external dependency becomes one of the internal dependencies that define this larger composite dependency structure. So the model could further track the external dependencies of the new larger composite dependency structure. This recursion predicts that there are latent dependency relations that cannot be observed directly in a dataset. An example is illustrated in (figure 11). To learn the outer dependency relation ABC-D, the model has to learn the inner dependency and second-level dependency first. These inner dependencies are the building blocks to the outer dependencies. The current experiments do not test recursion directly; they establish only the first step required by such process.

\begin{figure*}[!t]
\centering
\includegraphics[width=.55\textwidth,height=.31\textheight,keepaspectratio]{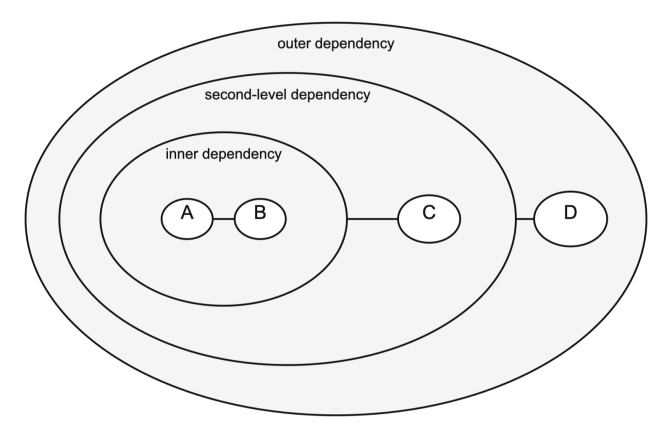}
\caption{recursion of internal-external dependency nesting}
\label{fig:11}
\end{figure*}
The meaning of such dependency recursion is nested event structures. For example, consider the question in (10). To generate the logical answer to this prompt, models need to draw on three levels of representations. First, the lexical semantics of tokens. Second, the semantics of the three grammatical structures in the three sentences. This corresponds to the first level of dependency nesting, which captures the event structures encoded in the grammatical structures. Third, models need to have representations for the global event structure, which captures the relations between the smaller event structures encoded in grammatical structures. In other words, to understand the whole text, models need to learn the semantic relations between these component sentences. The sentences, which were the composites in the first layer of nesting, become the components in the second layer of nesting. This hypothetical recursion might constitute a statistical learning approach to emergence, namely why tracking distributional statistics in large corpora could derive language intelligence. The basis of the reasoning ability of LLMs might be such event structures that capture the event structures beyond sentences. This prediction is not examined in the current study.

(10) ``The taller a person is, the more sophisticated that person is in linguistics.

Mary is taller than John.

So who is more sophisticated in linguistics?''

In summary, this study proposed a statistical learning mechanism of tracking the external distributions of composite dependency structures. The findings demonstrate that NLM statistical learning is not merely the accumulation of token-level co-occurrence statistics, but a process in which learned dependency structures become new distributional units for further learning. This statistical learning mechanism potentially helps NLMs solve the compositionality problem in the representation of grammar and potentially constitute a statistical learning account to emergence. It offers a new way of describing language representations in NLMs that is grounded in statistical structure rather than predefined linguistic categories. This perspective draws insight from statistical learning theory in human cognitive science, providing a common framework for understanding how abstract linguistic knowledge may emerge from statistical distributions.

\section{Materials and Methods}
\subsection{Synthetic grammar implementation}
The conceptual design and composition of the artificial languages are described in the main text. The implementation contained 180 ordinary lexical tokens: 20 tokens in each of the A, B, M, and N categories and 100 U tokens. Four additional identifiers represented the [PAD], [UNK], [CLS], and [MASK] tokens. Category labels were used only to generate the dataset and evaluate the trained models; they were not shown to the models during training. Before model input, the identifiers of the 180 ordinary tokens were randomly shuffled so that category membership could not be inferred from adjacent numerical identifiers.

For each dependency condition, the dataset generator enumerated all \(20^{4}=160{,}000\) possible combinations of the three lexical tokens in the relevant trigram and the associated M or N token. Corresponding BAA--M and ABA--N sequences were generated with random U-token backgrounds, trigram positions, and contextual-token positions. The compensation dataset was generated independently of the trigram conditions and contained controlled proportion of M--N, M-only, N-only, and U-only subsets. Automated checks verified the intended lexical balance of the M and N categories, \(P(M)=P(N)=0.40\), and \(P(M,N)=P(M)P(N)=0.16\).

The control dataset was generated using the same procedure. The only change was the replacement of the BAA trigrams by ABA trigrams, so that both M and N occurred in the context of the same ABA structure.

\subsection{Architecture and training for vector analysis models}
The main model contained 8 Transformer layers, 8 attention heads, a hidden dimension of 32, and an intermediate dimension of 64. The two scaling models were 4L/4H/16D with an intermediate dimension of 32 and 8L/8H/64D with an intermediate dimension of 128. All models used dynamic exact-two-position masking, hidden and attention dropout probabilities of 0.1, AdamW, a minibatch size of 64, a constant learning rate of \(1\times10^{-4}\), and weight decay of 0.01. For the reported developmental visualizations, checkpoints from steps 0--100,000 were retained every 200 updates, yielding 501 states including initialization. The displayed models used seed 1000. The architecture-matched ABA-only control model shown in Figure 2, 8, 10 used the same model and optimization settings for 1,500,000 updates. It was saved every 1,000 updates, yielding 1,501 states including initialization.

\subsection{Architecture and training for Probability-mass analysis models}
The probability-mass analysis used three separately trained architectures: 4L/4H/16D with intermediate dimension 32, 8L/8H/32D with intermediate dimension 64, and 8L/8H/64D with intermediate dimension 128. Training used dynamic exact-two-position masking, AdamW, a minibatch size of 512, a constant learning rate of \(5\times10^{-4}\), weight decay of 0.01, and hidden and attention dropout probabilities of 0.1. Each model was trained for 200,000 updates and saved at initialization and every 500 updates thereafter, yielding 401 states per model.

\subsection{Developmental Value-vector analysis}
Token vectors were extracted from every saved model state. At each checkpoint, every A, B, M, N, and U token identity was placed in the fourth position of a seven-token probe, with a fixed U token occupying the remaining positions. We extracted the Value projection,

\(V=XW_{V}+b_{V},\)

at the target position in the first Transformer layer. Because this projection is computed from the embedding-level input before the first contextual attention operation, it constitutes a static context-free representation. For the main model, each attention head contributed a four-dimensional vector. The 8 head-specific vectors were concatenated to produce one 32-dimensional vector for each token.

For each trained model, vectors from all token identities and all saved checkpoints were extracted. A separate autoencoder was then fitted for each model. For the main model, the autoencoder had the architecture \(32\to16\to2\to16\to32\), with GELU activations. It was optimized using AdamW, a learning rate of \(10^{-5}\), weight decay of \(10^{-4}\), a minibatch size of 512, and mean-squared reconstruction error. It is trained for half epoch, with no early stopping or drop out.

The two-dimensional autoencoder coordinates were used only for visualization. Because separate autoencoders were fitted to different models, the orientations, scales, and absolute distances of axes should not be compared across panels.

\subsection{Probability-mass evaluation}
The four probe families used to assess internal and external dependencies are described in the main text. Each probe retained metadata identifying its source sequence, motif location, masked position or positions, target category, and, for external-dependency probes, the target side. Probe strings were constructed from held-out sequences and were not used for model training.

For a masked position (j), logits were converted to probabilities using the softmax function. The probability mass assigned to target category (C) was calculated as

\(\operatorname{PM}(C\mid j)=\sum_{w\in C}P(w\mid j).\)

Thus, the measure evaluated the total probability assigned to the appropriate lexical category rather than the probability of a single token. At every checkpoint, 512 probe strings from each of the four probe families were evaluated. Probability mass was first averaged across probe strings within each model and checkpoint. These model-level means were then averaged across the 100 independently initialized models. Confidence intervals were calculated across model-level means, treating the model---not the individual probe string---as the independent unit.

\subsection{Scaling analysis and reproducibility}
For the scaling analysis, dataset generation, dynamic masking, optimization, checkpointing, probe construction, probability-mass calculation, and aggregation were held constant across architectures. Only the number of Transformer layers, number of attention heads, and hidden dimension were changed. Each architecture was trained under multiple random seeds, and model-level results were aggregated using the same procedure as for the main architecture.

\balance
\section*{References}
\small
\par\hangindent=1.5em\noindent Ahuja, K., Balachandran, V., Panwar, M., He, T., Smith, N. A., Goyal, N., \& Tsvetkov, Y. (2024). Learning syntax without planting trees: Understanding hierarchical generalization in transformers [Preprint]. arXiv. \url{https://doi.org/10.48550/arXiv.2404.16367}
\par\hangindent=1.5em\noindent Baker, M. C. (1997). Thematic roles and syntactic structure. In L. Haegeman (Ed.), Elements of grammar: Handbook in generative syntax (pp. 73--137). Kluwer Academic Publishers.
\par\hangindent=1.5em\noindent Bojun, W., Jenkins, H., \& Wonnacott, E. (2026). Developmental approach reveals the statistical learning of Neural Language Models: Transformers generalize from the most abstract statistical patterns.~arXiv preprint arXiv:2606.27460.
\par\hangindent=1.5em\noindent Bresnan, J., Cueni, A., Nikitina, T., \& Baayen, R. H. (2007). Predicting the dative alternation. In G. Bouma, I. Kr\"{a}mer, \& J. Zwarts (Eds.), Cognitive foundations of interpretation (pp. 69--94). Royal Netherlands Academy of Arts and Sciences.
\par\hangindent=1.5em\noindent Brown, H., Smith, K., Samara, A., \& Wonnacott, E. (2022). Semantic cues in language learning: An artificial language study with adult and child learners. Language, Cognition and Neuroscience, 37(4), 509--531. \url{https://doi.org/10.1080/23273798.2021.1995612}
\par\hangindent=1.5em\noindent Croft, W. (2001). Radical construction grammar: Syntactic theory in typological perspective. Oxford University Press. \url{https://doi.org/10.1093/acprof:oso/9780198299554.001.0001}
\par\hangindent=1.5em\noindent Croft, W. A. (2003). Lexical rules vs. constructions: A false dichotomy. In H. Cuyckens, T. Berg, R. Dirven, \& K.-U. Panther (Eds.), Current issues in linguistic theory (Vol. 243, pp. 49--68). John Benjamins Publishing Company. \url{https://doi.org/10.1075/cilt.243.07cro}
\par\hangindent=1.5em\noindent Croft, W. (2015). Force dynamics and directed change in event lexicalization and argument realization. In R. G. de Almeida \& C. Manouilidou (Eds.), Cognitive science perspectives on verb representation and processing (pp. 103--129). Springer International Publishing. \url{https://doi.org/10.1007/978-3-319-10112-5_5}
\par\hangindent=1.5em\noindent Croft, W., \& Cruse, D. A. (2004). Cognitive linguistics. Cambridge University Press.
\par\hangindent=1.5em\noindent Culicover, P. W., \& Jackendoff, R. (1999). The view from the periphery: The English comparative correlative. Linguistic Inquiry, 30(4), 543--571.
\par\hangindent=1.5em\noindent Dowty, D. R. (1991). Thematic proto-roles and argument selection. Language, 67(3), 547--619. \url{https://doi.org/10.1353/lan.1991.0021}
\par\hangindent=1.5em\noindent Futrell, R., \& Mahowald, K. (2025). How linguistics learned to stop worrying and love the language models. Behavioral and Brain Sciences, 1--98. \url{https://doi.org/10.1017/S0140525X2510112X}
\par\hangindent=1.5em\noindent Futrell, R., Wilcox, E., Morita, T., Qian, P., Ballesteros, M., \& Levy, R. (2019). Neural language models as psycholinguistic subjects: Representations of syntactic state [Preprint]. arXiv. \url{https://doi.org/10.48550/arXiv.1903.03260}
\par\hangindent=1.5em\noindent Goldberg, A. E. (1995). Constructions: A construction grammar approach to argument structure. University of Chicago Press. \url{https://press.uchicago.edu/ucp/books/book/chicago/C/bo3683810.html}
\par\hangindent=1.5em\noindent Goldberg, A. E. (2006). Constructions at work: The nature of generalization in language. Oxford University Press. \url{https://doi.org/10.1093/acprof:oso/9780199268511.001.0001}
\par\hangindent=1.5em\noindent Goldberg, A. E. (2019). Explain me this: Creativity, competition, and the partial productivity of constructions. Princeton University Press. \url{https://doi.org/10.2307/j.ctvc772nn}
\par\hangindent=1.5em\noindent Hardy, M., Sucholutsky, I., Thompson, B., \& Griffiths, T. (2023). Large language models meet cognitive science: LLMs as tools, models, and participants. Proceedings of the Annual Meeting of the Cognitive Science Society, 45(45), 14--15. \url{https://escholarship.org/uc/item/6dp9k2gz}
\par\hangindent=1.5em\noindent Hewitt, J., \& Manning, C. D. (2019). A structural probe for finding syntax in word representations. In J. Burstein, C. Doran, \& T. Solorio (Eds.), Proceedings of the 2019 Conference of the North American Chapter of the Association for Computational Linguistics: Human Language Technologies, Volume 1 (Long and Short Papers) (pp. 4129--4138). Association for Computational Linguistics. \url{https://doi.org/10.18653/v1/N19-1419}
\par\hangindent=1.5em\noindent Hilpert, M. (2019). Construction grammar and its application to English. Edinburgh University Press. \url{https://doi.org/10.1515/9781474433624}
\par\hangindent=1.5em\noindent Hofmann, V., Weissweiler, L., Mortensen, D. R., Sch\"{u}tze, H., \& Pierrehumbert, J. B. (2025). Derivational morphology reveals analogical generalization in large language models. Proceedings of the National Academy of Sciences, 122(19), e2423232122. \url{https://doi.org/10.1073/pnas.2423232122}
\par\hangindent=1.5em\noindent Huang, J., \& Chang, K. C.-C. (2023, July). Towards reasoning in large language models: A survey. In Findings of the Association for Computational Linguistics: ACL 2023 (pp. 1049--1065). Association for Computational Linguistics. \url{https://doi.org/10.18653/v1/2023.findings-acl.67}
\par\hangindent=1.5em\noindent Hupkes, D., Dankers, V., Mul, M., \& Bruni, E. (2020). Compositionality decomposed: How do neural networks generalise? Journal of Artificial Intelligence Research, 67, 757--795. \url{https://doi.org/10.1613/jair.1.11674}
\par\hangindent=1.5em\noindent Isbilen, E. S., \& Christiansen, M. H. (2022). Statistical learning of language: A meta-analysis into 25 years of research. Cognitive Science, 46(9), e13198. \url{https://doi.org/10.1111/cogs.13198}
\par\hangindent=1.5em\noindent Kallens, P., Kristensen-McLachlan, R. D., \& Christiansen, M. H. (2023). Large language models demonstrate the potential of statistical learning in language. Cognitive Science, 47(3), e13256. \url{https://doi.org/10.1111/cogs.13256}
\par\hangindent=1.5em\noindent Kapatsinski, V. (2026). Transformers perform adaptive partial pooling. Proceedings of the Annual Meeting of the Cognitive Science Society, 48, 3050--3057. \url{https://doi.org/10.48550/arXiv.2602.03980}
\par\hangindent=1.5em\noindent Kim, N., \& Smolensky, P. (2021). Testing for grammatical category abstraction in neural language models. In A. Ettinger, E. Pavlick, \& B. Prickett (Eds.), Proceedings of the Society for Computation in Linguistics 2021 (pp. 467--470). Association for Computational Linguistics. \url{https://aclanthology.org/2021.scil-1.59/}
\par\hangindent=1.5em\noindent Kim, N., \& Linzen, T. (2020). COGS: A compositional generalization challenge based on semantic interpretation. In Proceedings of EMNLP 2020 (pp. 9087--9105). \url{https://doi.org/10.18653/v1/2020.emnlp-main.731}
\par\hangindent=1.5em\noindent Lakoff, G. (1990). The invariance hypothesis: Is abstract reason based on image-schemas? Cognitive Linguistics, 1(1), 39--74. \url{https://doi.org/10.1515/cogl.1990.1.1.39}
\par\hangindent=1.5em\noindent Lake, B. M., \& Baroni, M. (2018). Generalization without systematicity: On the compositional skills of sequence-to-sequence recurrent networks. In Proceedings of ICML 2018 (pp. 2873--2882). \href{https://proceedings.mlr.press/v80/lake18a.html}{PMLR}.
\par\hangindent=1.5em\noindent Lakretz, Y., Hupkes, D., Vergallito, A., Marelli, M., Baroni, M., \& Dehaene, S. (2021). Mechanisms for handling nested dependencies in neural-network language models and humans. Cognition, 213, 104699. \url{https://doi.org/10.1016/j.cognition.2021.104699}
\par\hangindent=1.5em\noindent Langacker, R. W. (1987). Foundations of cognitive grammar: Volume I: Theoretical prerequisites. Stanford University Press.
\par\hangindent=1.5em\noindent Langacker, R. W. (2009). Investigations in cognitive grammar. Walter de Gruyter.
\par\hangindent=1.5em\noindent Lany, J., \& Saffran, J. R. (2010). From statistics to meaning: Infants' acquisition of lexical categories. Psychological Science, 21(2), 284--291. \url{https://doi.org/10.1177/0956797609358570}
\par\hangindent=1.5em\noindent Lany, J., \& Saffran, J. R. (2011). Interactions between statistical and semantic information in infant language development. Developmental Science, 14(5), 1207--1219. \url{https://doi.org/10.1111/j.1467-7687.2011.01073.x}
\par\hangindent=1.5em\noindent Levin, B. (2015). Semantics and pragmatics of argument alternations. Annual Review of Linguistics, 1, 63--83. \url{https://doi.org/10.1146/annurev-linguist-030514-125141}
\par\hangindent=1.5em\noindent Levin, B., \& Rappaport Hovav, M. (2005). Argument realization. Cambridge University Press. \url{https://doi.org/10.1017/CBO9780511610479}
\par\hangindent=1.5em\noindent Li, J., \& Liu, Y. (2025). An investigation of comparative correlative constructions in auto-regressive large language models: From construction grammar to computational understanding [Preprint]. Research Square. \url{https://doi.org/10.21203/rs.3.rs-6702743/v1}
\par\hangindent=1.5em\noindent Linzen, T., Dupoux, E., \& Goldberg, Y. (2016). Assessing the ability of LSTMs to learn syntax-sensitive dependencies. Transactions of the Association for Computational Linguistics, 4, 521--535. \url{https://doi.org/10.1162/tacl_a_00115}
\par\hangindent=1.5em\noindent Marro, S., Evangelista, D., Huang, X. A., La Malfa, E., Lombardi, M., \& Wooldridge, M. J. (2025). Language models are implicitly continuous. In The Thirteenth International Conference on Learning Representations. \url{https://openreview.net/forum?id=SMK0f8JoKF}
\par\hangindent=1.5em\noindent Mintz, T. H. (2002). Category induction from distributional cues in an artificial language. Memory \& Cognition, 30(5), 678--686. \url{https://doi.org/10.3758/BF03196424}
\par\hangindent=1.5em\noindent Misyak, J. B., Christiansen, M. H., \& Tomblin, J. B. (2009). Statistical learning of nonadjacencies predicts on-line processing of long-distance dependencies in natural language. In N. Taatgen, H. van Rijn, J. Nerbonne, \& L. Schomaker (Eds.), Proceedings of the 31st Annual Conference of the Cognitive Science Society (pp. 177--182). Cognitive Science Society.
\par\hangindent=1.5em\noindent Mondorf, P., \& Plank, B. (2024). Beyond accuracy: Evaluating the reasoning behavior of large language models---A survey. In Proceedings of the First Conference on Language Modeling. \url{https://openreview.net/forum?id=Lmjgl2n11u}
\par\hangindent=1.5em\noindent Morgan, J. L., \& Newport, E. L. (1981). The role of constituent structure in the induction of an artificial language. Journal of Verbal Learning and Verbal Behavior, 20(1), 67--85. \url{https://doi.org/10.1016/S0022-5371(81)90312-1}
\par\hangindent=1.5em\noindent Murty, S., Sharma, P., Andreas, J., \& Manning, C. D. (2023). Grokking of hierarchical structure in vanilla transformers [Preprint]. arXiv. \url{https://doi.org/10.48550/arXiv.2305.18741}
\par\hangindent=1.5em\noindent Nunberg, G., Sag, I. A., \& Wasow, T. (1994). Idioms. Language, 70(3), 491--538. \url{https://doi.org/10.2307/416483}
\par\hangindent=1.5em\noindent Patil, A., \& Jadon, A. (2025). Advancing reasoning in large language models: Promising methods and approaches [Preprint]. arXiv. \url{https://doi.org/10.48550/arXiv.2502.03671}
\par\hangindent=1.5em\noindent Pelucchi, B., Hay, J. F., \& Saffran, J. R. (2009). Learning in reverse: Eight-month-old infants track backward transitional probabilities. Cognition, 113(2), 244--247. \url{https://doi.org/10.1016/j.cognition.2009.07.011}
\par\hangindent=1.5em\noindent Perek, F. (2015). Argument structure in usage-based construction grammar: Experimental and corpus-based perspectives (Vol. 17). John Benjamins Publishing Company. \url{https://doi.org/10.1075/cal.17}
\par\hangindent=1.5em\noindent Perek, F., \& Goldberg, A. E. (2015). Generalizing beyond the input: The functions of the constructions matter. Journal of Memory and Language, 84, 108--127. \url{https://doi.org/10.1016/j.jml.2015.04.006}
\par\hangindent=1.5em\noindent Perek, F., \& Goldberg, A. E. (2017). Linguistic generalization on the basis of function and constraints on the basis of statistical preemption. Cognition, 168, 276--293. \url{https://doi.org/10.1016/j.cognition.2017.06.019}
\par\hangindent=1.5em\noindent Petersen, E., \& Potts, C. (2023, May). Lexical semantics with large language models: A case study of English ``break''. In Findings of the Association for Computational Linguistics: EACL 2023 (pp. 490--511). Association for Computational Linguistics. \url{https://doi.org/10.18653/v1/2023.findings-eacl.36}
\par\hangindent=1.5em\noindent Pinker, S. (1989). Learnability and cognition: The acquisition of argument structure. The MIT Press.
\par\hangindent=1.5em\noindent Rappaport Hovav, M., \& Levin, B. (2008). The English dative alternation: The case for verb sensitivity. Journal of Linguistics, 44(1), 129--167. \url{https://doi.org/10.1017/S0022226707004975}
\par\hangindent=1.5em\noindent Reeder, P. A., Newport, E. L., \& Aslin, R. N. (2013). From shared contexts to syntactic categories: The role of distributional information in learning linguistic form-classes. Cognitive Psychology, 66(1), 30--54. \url{https://doi.org/10.1016/j.cogpsych.2012.09.001}
\par\hangindent=1.5em\noindent Reeder, P. A., Newport, E. L., \& Aslin, R. N. (2017). Distributional learning of subcategories in an artificial grammar: Category generalization and subcategory restrictions. Journal of Memory and Language, 97, 17--29. \url{https://doi.org/10.1016/j.jml.2017.07.006}
\par\hangindent=1.5em\noindent Romberg, A. R., \& Saffran, J. R. (2010). Statistical learning and language acquisition. WIREs Cognitive Science, 1(6), 906--914. \url{https://doi.org/10.1002/wcs.78}
\par\hangindent=1.5em\noindent Saffran, J. R. (2001). The use of predictive dependencies in language learning. Journal of Memory and Language, 44(4), 493--515. \url{https://doi.org/10.1006/jmla.2000.2759}
\par\hangindent=1.5em\noindent Saffran, J. R. (2020). Statistical language learning in infancy. Child Development Perspectives, 14(1), 49--54. \url{https://doi.org/10.1111/cdep.12355}
\par\hangindent=1.5em\noindent Saffran, J. R., Aslin, R. N., \& Newport, E. L. (1996). Statistical learning by 8-month-old infants. Science, 274(5294), 1926--1928. \url{https://doi.org/10.1126/science.274.5294.1926}
\par\hangindent=1.5em\noindent Samara, A., Wonnacott, E., Saxena, G., Maitreyee, R., Fazekas, J., \& Ambridge, B. (2025). Learners restrict their linguistic generalizations using preemption but not entrenchment: Evidence from artificial-language-learning studies with adults and children. Psychological Review, 132(1), 1--17. \url{https://doi.org/10.1037/rev0000463}
\par\hangindent=1.5em\noindent Smith, K. H. (1969). Learning co-occurrence restrictions: Rule induction or rote learning? Journal of Verbal Learning and Verbal Behavior, 8(2), 319--321. \url{https://doi.org/10.1016/S0022-5371(69)80086-1}
\par\hangindent=1.5em\noindent Theijssen, D., ten Bosch, L., Boves, L., Cranen, B., \& van Halteren, H. (2013). Choosing alternatives: Using Bayesian networks and memory-based learning to study the dative alternation. Corpus Linguistics and Linguistic Theory, 9(2), 227--262. \url{https://doi.org/10.1515/cllt-2013-0007}
\par\hangindent=1.5em\noindent Thompson, S. P., \& Newport, E. L. (2007). Statistical learning of syntax: The role of transitional probability. Language Learning and Development, 3(1), 1--42.
\par\hangindent=1.5em\noindent Tomasello, M. (2003). Constructing a language: A usage-based theory of language acquisition. Harvard University Press. \url{https://doi.org/10.2307/j.ctv26070v8}
\par\hangindent=1.5em\noindent Tomasello, M. (2007). Acquiring linguistic constructions. In W. Damon \& R. M. Lerner (Eds.), Handbook of child psychology (1st ed.). Wiley. \url{https://doi.org/10.1002/9780470147658.chpsy0206}
\par\hangindent=1.5em\noindent V\'{a}zquez Mart\'{i}nez, H. J., Heuser, A., Yang, C., \& Kodner, J. (2023, December). Evaluating neural language models as cognitive models of language acquisition. In~Proceedings of the 1st GenBench Workshop on (Benchmarking) Generalisation in NLP~(pp. 48-64).
\par\hangindent=1.5em\noindent Vuli\'{c}, I., Ponti, E. M., Litschko, R., Glava\v{s}, G., \& Korhonen, A. (2020, November). Probing pretrained language models for lexical semantics. In Proceedings of the 2020 Conference on Empirical Methods in Natural Language Processing (EMNLP) (pp. 7222--7240). Association for Computational Linguistics. \url{https://doi.org/10.18653/v1/2020.emnlp-main.586}
\par\hangindent=1.5em\noindent Wei, J., Garrette, D., Linzen, T., \& Pavlick, E. (2021). Frequency effects on syntactic rule learning in transformers [Preprint]. arXiv. \url{https://doi.org/10.48550/arXiv.2109.07020}
\par\hangindent=1.5em\noindent Weissweiler, L., He, T., Otani, N., Mortensen, D. R., Levin, L., \& Sch\"{u}tze, H. (2023a). Construction grammar provides unique insight into neural language models [Preprint]. arXiv. \url{https://doi.org/10.48550/arXiv.2302.02178}
\par\hangindent=1.5em\noindent Weissweiler, L., Hofmann, V., K\"{o}ksal, A., \& Sch\"{u}tze, H. (2023b). Explaining pretrained language models' understanding of linguistic structures using construction grammar. Frontiers in Artificial Intelligence, 6, 1225791. \url{https://doi.org/10.3389/frai.2023.1225791}
\par\hangindent=1.5em\noindent Wilcox, E. G., Futrell, R., \& Levy, R. (2022). Using computational models to test syntactic learnability. Linguistic Inquiry. 55(4), 805-848. \url{https://doi.org/10.1162/ling_a_00491}
\par\hangindent=1.5em\noindent Wonnacott, E. (2013). Learning: Statistical mechanisms in language acquisition. In P.-M. Binder \& K. Smith (Eds.), The language phenomenon (pp. 65--92). Springer Berlin Heidelberg. \url{https://doi.org/10.1007/978-3-642-36086-2_4}
\par\hangindent=1.5em\noindent Wonnacott, E., Brown, H., \& Nation, K. (2017). Skewing the evidence: The effect of input structure on child and adult learning of lexically based patterns in an artificial language. Journal of Memory and Language, 95, 36--48. \url{https://doi.org/10.1016/j.jml.2017.01.005}
\end{document}